%% file: main.tex
\documentclass[letterpaper, 10pt, conference]{ieeeconf}

\IEEEoverridecommandlockouts   %
\newcommand{\citep}[2][]{\ifx\\#1\\\cite{#2}\else\cite[#1]{#2}\fi}
\usepackage{xcolor}
\usepackage{amsmath, amssymb, amsfonts}
\usepackage{bm}
\makeatletter
\let\NAT@parse\undefined
\makeatother
\usepackage{cite}       %
\usepackage[hidelinks]{hyperref}
\hypersetup{pdfauthor={Zhewei Gui, Anthony Bolton, Jiyao Zhang, Zeyuan Chen, Qiyang Yan, Tianyu Wang, Hao Dong},pdftitle={DexHOT: Learning Hand--Object Configuration Transfer for Dexterous Manipulation}}
\usepackage{graphicx}
\usepackage{multirow}    %
\usepackage{booktabs}    %
\usepackage{tabularx}    %
\makeatletter
\long\def\dexhot@setcaption#1{%
  \setbox\@tempboxa\hbox{\footnotesize\normalfont #1}%
  \ifdim\wd\@tempboxa>\hsize
    {\footnotesize\normalfont\noindent #1\par}%
  \else
    \hbox to\hsize{\hfil\box\@tempboxa\hfil}%
  \fi}
\long\def\@makecaption#1#2{%
  \ifx\@captype\@IEEEtablestring
    \dexhot@setcaption{#1: #2}\vskip 6pt
  \else
    \@IEEEfigurecaptionsepspace
    \dexhot@setcaption{#1.~~ #2}%
  \fi}
\newlength{\dexhot@topfloatshift}
\def\dexhot@tfshift{\ifnum\c@page>\@ne \dexhot@topfloatshift\else\z@\fi}
\def \@cflt{%
    \let \@elt \@comflelt
    \setbox\@tempboxa \vbox{}%
    \@toplist
    \setbox\@outputbox \vbox{%
                             \boxmaxdepth \maxdepth
                             \vskip \dexhot@tfshift
                             \unvbox\@tempboxa
                             \vskip -\floatsep
                             \topfigrule
                             \vskip \textfloatsep
                             \vskip -\dexhot@tfshift
                             \unvbox\@outputbox
                             }%
    \let\@elt\relax
    \xdef\@freelist{\@freelist\@toplist}%
    \global\let\@toplist\@empty
}
\def \@combinedblfloats{%
  \ifx \@dbltoplist \@empty
  \else
    \setbox\@tempboxa \vbox{}%
    \let \@elt \@comdblflelt
    \@dbltoplist
    \let \@elt \relax
    \xdef \@freelist {\@freelist\@dbltoplist}%
    \global\let \@dbltoplist \@empty
    \setbox\@outputbox \vbox to\textheight
      {\vskip \dexhot@tfshift
       \unvbox\@tempboxa\vskip-\dblfloatsep
       \ifnum \@dbltopnum>\m@ne
         \dblfigrule
       \fi
       \vskip \dbltextfloatsep
       \vskip -\dexhot@tfshift
       \unvbox\@outputbox
       }%
  \fi
}
\def\subsubsection{\@startsection{subsubsection}{3}{\z@}{0ex plus 0.1ex minus 0.1ex}%
{0ex}{\normalfont\normalsize\bfseries}}
\def\@IEEEsectpunct{.\ \,}
\def\@IEEEauthorblockconfadjspace{-0.85em}
\def\@IEEEauthorblockAtopspace{4pt}
\makeatother
\graphicspath{{Figures/}}

\title{\LARGE \bf
DexHOT: Learning Hand--Object Configuration Transfer for Dexterous Manipulation
}

\newcommand{\authornotemark}[1]{\textsuperscript{\sublargesize\ensuremath{#1}}}
\newcommand{\authoraffilmark}[1]{\textsuperscript{\sublargesize #1}}
\author{%
\authorblockN{%
\bfseries
Zhewei Gui$^{1,3,*}$\quad Anthony Bolton$^{1,3,*}$\quad
Jiyao Zhang$^{1,2,3,\ddagger}$\\[2pt]
Zeyuan Chen$^{1,4}$\quad Qiyang Yan$^{1}$\quad
Tianyu Wang$^{1,4}$\quad Hao Dong$^{1,2,3,\dagger}$}
\authorblockA{%
\sublargesize \authoraffilmark{1}CFCS, School of CS, PKU, China\\
\sublargesize \authoraffilmark{2}National Key Laboratory for Multimedia Information Processing, School of CS, PKU, China\\
\sublargesize \authoraffilmark{3}PrimeBot\qquad \authoraffilmark{4}State Key Laboratory of General Artificial Intelligence, BIGAI\\[2pt]
\sublargesize \authornotemark{*}Co-first Authors\quad \authornotemark{\dagger}Corresponding author\quad \authornotemark{\ddagger}Project Lead\\[3pt]
\normalsize \href{https://dex-hot.github.io/}{\textcolor{magenta}{\texttt{https://dex-hot.github.io/}}}}%
}

\IEEEaftertitletext{\input{Figures/teaser}}

\begin{document}

\maketitle
\thispagestyle{empty}
\pagestyle{empty}

\input{Texs/abstract}

\input{Texs/introduction}

\input{Texs/related_work}

\input{Texs/problem_formulation}
\input{Texs/method}
\input{Texs/experiments}
\input{Texs/conclusion}

\section*{ACKNOWLEDGMENT}
We thank Tianhao Wu, Zihan Ding, Junhan Wang, Zimu Han and Yiming Zeng for their insightful discussions,
and Hangqi Zhou and Zihao Zhao for their assistance with the hardware. This research was
supported by Beijing Natural Science Foundation (26L080330) and National
Natural Science Foundation of China (62376006).

\bibliographystyle{IEEEtran}   %
\bibliography{example}  %

\end{document}

%% file: Figures/teaser.tex
\begin{minipage}{\textwidth}
\centering
\includegraphics[width=\textwidth,pagebox=cropbox]{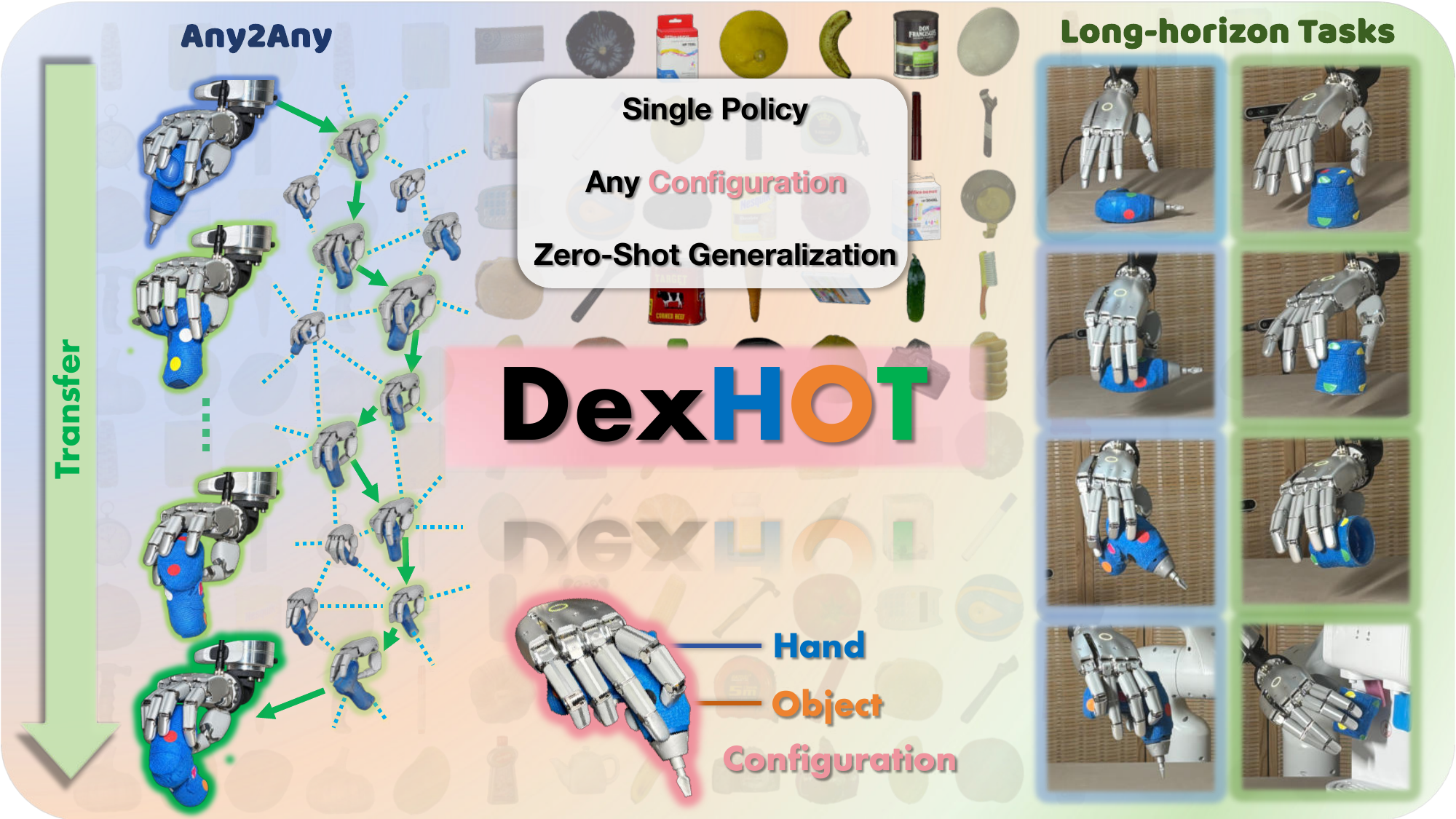}
\makeatletter
\def\@captype{figure}
\makeatother
\vspace{-12pt}
\caption{\textbf{DexHOT} learns a single policy across multiple objects to transfer between configurations that jointly specify hand articulation and object pose relative to the wrist.
\emph{Left:} a greedy planner guides transfer to a distant target through subgoals selected from a synthetic configuration pool.
\emph{Right:} human demonstrations provide subgoals for long-horizon tasks involving previously unseen hand--object configurations.}
\label{fig:teaser}
\end{minipage}

%% file: Texs/abstract.tex
\begin{abstract}
Dexterous manipulation involves not only moving objects, but also arranging the hand around them to maintain useful contacts and enable subsequent actions. Yet many learning-based systems specify goals primarily through object poses or trajectories, leaving hand--object configurations to emerge implicitly. We introduce \textbf{DexHOT}, which formulates dexterous manipulation as \emph{hand--object configuration transfer}: moving between feasible configurations that jointly specify hand articulation and object pose relative to the wrist. DexHOT learns a reusable relation-conditioned transfer primitive from synthetic feasible configuration pairs, without human reference motions, and composes long-horizon behaviors through closed-loop tracking of configuration subgoals. On DexHOTBench with 120 objects, a single policy achieves 80.5\% transfer success on seen objects and 69.4\% on unseen objects; on a real dexterous hand, it achieves 61.5\% success over 200 real-robot transfers and 90\% task success across six demonstration-guided tasks. These results show that explicit hand--object goals provide an expressive and composable interface for dexterous manipulation.
More details and videos are available at \href{https://dex-hot.github.io/}{\texttt{https://dex-hot.github.io/}}.
\end{abstract}

%% file: Texs/introduction.tex
\section{Introduction}
\label{sec:introduction}

Dexterous manipulation depends on both object motion and hand arrangement.
A mug held by its rim, handle, or body can have the same pose, yet differ in stability, functionality, and subsequent manipulation options.
Object-state goals cannot express these differences.

Recent learned systems span increasingly complex capabilities, from dexterous grasping~\citep{clutterdexgrasp,graspgf,cadgrasp} and task-specific in-hand reorientation~\citep{openai_dactyl,openai_rubiks,visual_dexterity,hora,rotateit,drop,simple2complex} to tool manipulation~\citep{simtoolreal} and more general dexterous skills~\citep{dexteritygen,wu2025canonical3dtactile,li2025adaptivevisuotactile}, together with robust sim-to-real transfer under challenging contact dynamics~\citep{dextreme,anyrotate,touch_dexterity,dexndm,dexpoint}.
Many use target object orientations or pose trajectories.
These compact goals make learning tractable but leave the hand--object relation implicit: hand configurations reaching the same object goal can differ substantially in usefulness for subsequent interaction.

This limitation becomes particularly important in long-horizon manipulation.
Regrasping, tool use, and functional interaction often require intermediate hand--object configurations that enable the next phase of the task~\citep{opendexgrasp,long2025checkmanual,zeng2024lvdiffusor,long2026scalingmanual}.
Contact-switching behaviors such as finger gaiting~\citep{han_gaiting} can emerge in object-goal systems~\citep{openai_dactyl,hora,dexndm}, but the goal does not explicitly prescribe the resulting hand configuration, and discovering these behaviors through exploration remains challenging~\citep{maniprrt}.

We therefore formulate dexterous manipulation as \emph{hand--object configuration transfer}.
A configuration specifies the hand articulation together with the object pose relative to the wrist, and a transfer seeks to move from one feasible configuration to another.
This changes the unit of goal-conditioned control from an object state to a hand--object relation, allowing both object motion and the resulting hand configuration to be specified explicitly.

Based on this formulation, we present \textbf{DexHOT}, a framework for learning reusable hand--object configuration transfer (Fig.~\ref{fig:teaser}).
DexHOT is built on the idea that long-range in-hand transfers can be composed from short, locally reachable transfers.
We synthesize diverse pools of physically feasible configurations, train a single relation-conditioned controller using local-goal sampling and population-based reinforcement learning (RL), and reuse it through hierarchical chaining of configuration subgoals.
The resulting controller exhibits dexterous skills such as finger gaiting and finger pivoting without planner-guided rollouts, an explicit exploration manifold, or skill-specific supervision during training.

To evaluate configuration transfer directly, we introduce \textbf{DexHOTBench}, comprising 120 objects with physically validated configuration pools and fixed transfer pairs spanning different distances.
The 80 held-out objects contain 40 unseen objects from training categories and 40 objects from unseen categories.
A single policy achieves 73.1\% success averaged over close and far transfers of all 120 objects, with 80.5\% on seen and 69.4\% on held-out objects, and 61.5\% success over 200 real-robot transfers spanning seen and unseen objects.

Our contributions are threefold:
\begin{itemize}
    \item We introduce \textbf{Hand--Object Configuration Transfer (HOT)}, a dexterous manipulation formulation in which goals explicitly specify both hand articulation and the relative hand--object pose.
\item We develop \textbf{DexHOT}, which learns a reusable configuration-transfer controller across objects from local feasible transfers and composes it into long-range and long-horizon dexterous behaviors. DexHOT exhibits emergent finger gaiting and finger pivoting and outperforms prior methods on demonstration-guided tasks.
    \item We introduce \textbf{DexHOTBench}, a 120-object benchmark for configuration transfer, and demonstrate generalization to held-out objects together with zero-shot transfer to real-robot manipulation.
\end{itemize}

%% file: Texs/related_work.tex
\section{Related Work}
\label{sec:related-work}

\subsection{Dexterous Manipulation and Goal Representations}

Large-scale sim-to-real reinforcement learning has enabled robust in-hand object reorientation on real dexterous hands~\citep{openai_dactyl,openai_rubiks,dextreme,visual_dexterity}.
Palm-up manipulation mitigates object dropping, a common failure mode in in-hand manipulation, whereas downward-facing manipulation is more challenging because gravity increases the risk of object loss~\citep{visual_dexterity,anyrotate,dexndm}.
DexHOT focuses on palm-down transfer, where the hand must actively maintain the object against gravity.
Prior work has also explored point-cloud policies for object generalization~\citep{dexpoint} and sampling-based predictive control~\citep{drop}.
SimToolReal~\citep{simtoolreal} learns a shared tool-manipulation policy conditioned on object-pose goals and a coarse grasp-region bounding box, but it does not explicitly prescribe the target hand articulation.
By contrast, DexHOT uses hand--object configurations that jointly specify hand articulation and the object's pose relative to the wrist.

DexterityGen~\citep{dexteritygen} and UniDexFPM~\citep{unidexfpm} also reason at the level of hand--object configurations.
DexterityGen uses a low-level controller to refine externally provided motion, whereas UniDexFPM provides a closed-loop interface for functional pre-grasp manipulation using tabletop support.
DexHOT instead learns configuration transfer directly and can follow planner-generated configuration subgoals in palm-down manipulation without a continuous human motion reference.

Geometry-aware curricula and goal-conditioned grasp proposals improve generalizable dexterous grasp acquisition~\citep{unidexgraspplusplus,unidexgrasp,anydexgrasp}. Population-based training improves exploration in hand--arm manipulation, including regrasping and reorientation~\citep{dexpbt}. DexHOT focuses on a reusable hand--object configuration interface for continued in-hand manipulation.

\subsection{Hand--Object Interaction Modeling}

DexTrack~\citep{dextrack} learns a generalizable controller for tracking human--object reference motions.
DexRepNet and DexRepNet++ explicitly encode hand--object geometry and spatial relations for dexterous manipulation~\citep{dexrepnet,dexrepnetpp}.
Interaction-aware methods such as DexHandDiff~\citep{dexhanddiff} likewise model contact structure to avoid physically inconsistent object motions, while FlowHOI and PKDA generate or transfer hand--object interaction trajectories from human-centric inputs~\citep{flowhoi,pkda}.
TeleDexter~\citep{teledexter} also uses hand--object co-tracking subgoals and trains object-specific controllers from human reference motions. DexHOT instead learns a shared transfer policy across objects from synthetic feasible configuration pairs, without demonstrations during policy training. At deployment, demonstrations provide the approach, wrist trajectory, and configuration goals.

%% file: Texs/problem_formulation.tex
\section{Problem Formulation}
\label{sec:formulation}

We formulate dexterous manipulation as hand--object configuration transfer: moving from an initial configuration to a target that specifies both the hand articulation and the object pose relative to the wrist. This makes the desired arrangement of the hand on the object an explicit control objective and provides a reusable goal interface for manipulation.

\subsection{Hand--Object Configuration}
Let $W$ and $O$ denote the wrist and object frames. We describe their kinematic configuration by
\begin{equation}
c = \bigl(\bm{q}_h,\ {}^{W}\!T_O\bigr)\in\mathbb{R}^{n_h}\times SE(3),
\label{eq:configuration}
\end{equation}
where $\bm{q}_h \in \mathbb{R}^{n_h}$ denotes the hand joint angles and ${}^{W}\!T_O=({}^{W}\!R_O,\,{}^{W}\!\bm{p}_O)$ is the object pose in the wrist frame. A target $c^\star$ specifies both the desired hand articulation and the desired position and orientation of the object relative to the wrist.

To couple hand articulation and relative object pose in a common geometric representation, we map $c$ to $K=28$ hand keypoints in the object frame: the palm reference point, the origins of the 22 finger-link frames, and five fingertip sites. With ${}^{W}\!\bm{x}_j(\bm{q}_h)$ denoting the forward-kinematics position of keypoint $j$ in the wrist frame, its object-frame position is
\begin{equation}
\bm{k}_j(c)
= {}^{W}\!R_O^{\mathsf T}
  \bigl({}^{W}\!\bm{x}_j(\bm{q}_h)-{}^{W}\!\bm{p}_O\bigr),
\label{eq:object-frame-keypoint}
\end{equation}
The ordered keypoints $\bm{k}(c)=[\bm{k}_1(c),\ldots,\bm{k}_K(c)]\in\mathbb{R}^{3\times K}$ capture the combined spatial effect of the hand joint angles and the object pose relative to the wrist. They are kinematic reference points, not assumed contact locations.

\subsection{Configuration Distance}
\label{sec:keypoint}
We compare two configurations by their mean keypoint displacement:
\begin{equation}
d(c,c')
= \frac{1}{K}\sum_{j=1}^{K}
  \left\lVert \bm{k}_j(c)-\bm{k}_j(c')\right\rVert_2 .
\label{eq:keypoint}
\end{equation}
This distance is invariant to rigid motion of the hand--object pair in the world. We report $d$ in millimeters.

\subsection{Configuration Transfer}
A transfer on object $o$ starts from a feasible configuration $c_0$ and aims to reach a feasible target $c^\star$ while keeping the object in the hand. At control step $t$, we write the recurrent transfer policy compactly as
\begin{equation}
\bm{a}_t
= \pi_\theta\!\left(c_t,\bm{s}_t,\bm{\phi}_o,c_t^\star;\bm{h}_t\right),
\label{eq:policy-interface}
\end{equation}
where $c_t$ is the current hand--object configuration, $\bm{s}_t$ denotes proprioceptive and control state, $\bm{\phi}_o$ is a coarse object-shape descriptor, $c_t^\star$ is the active target configuration, and $\bm{h}_t$ is the recurrent hidden state. The policy receives observable estimates and their feature encodings (Sec.~\ref{sec:obs}) at every control step, while $\bm{h}_t$ carries the preceding observation--action history. For a direct transfer, $c_t^\star=c^\star$ remains fixed; during hierarchical chaining, it denotes the currently active subgoal.

We learn the controller across objects using goal-conditioned reinforcement learning; the reward and training procedure are described in Sec.~\ref{sec:train}.

For evaluation, a transfer is successful if $d(c_t,c^\star)\leq\tau$ holds for $H=10$ consecutive control steps within the prescribed time limit, without dropping the object before completion. We use $\tau=20$\,mm and report the fraction of successful transfer trials. The formulation specifies the desired hand--object relation, while the learned controller determines the intermediate hand motions, finger gaits, and regrasps during closed-loop execution. This definition concerns in-hand transfer; the demonstration-guided experiments additionally reuse the same policy for pickup after a demonstration-driven pre-grasp approach (Sec.~\ref{sec:exp-real-hoi}).

%% file: Texs/method.tex
\section{DexHOT}
\label{sec:method}
Our key insight is to learn short transfers between nearby feasible hand--object configurations and compose them into longer manipulation behaviors. Such local transfers are easier to discover through exploration, while still requiring nontrivial changes in hand articulation and contact.

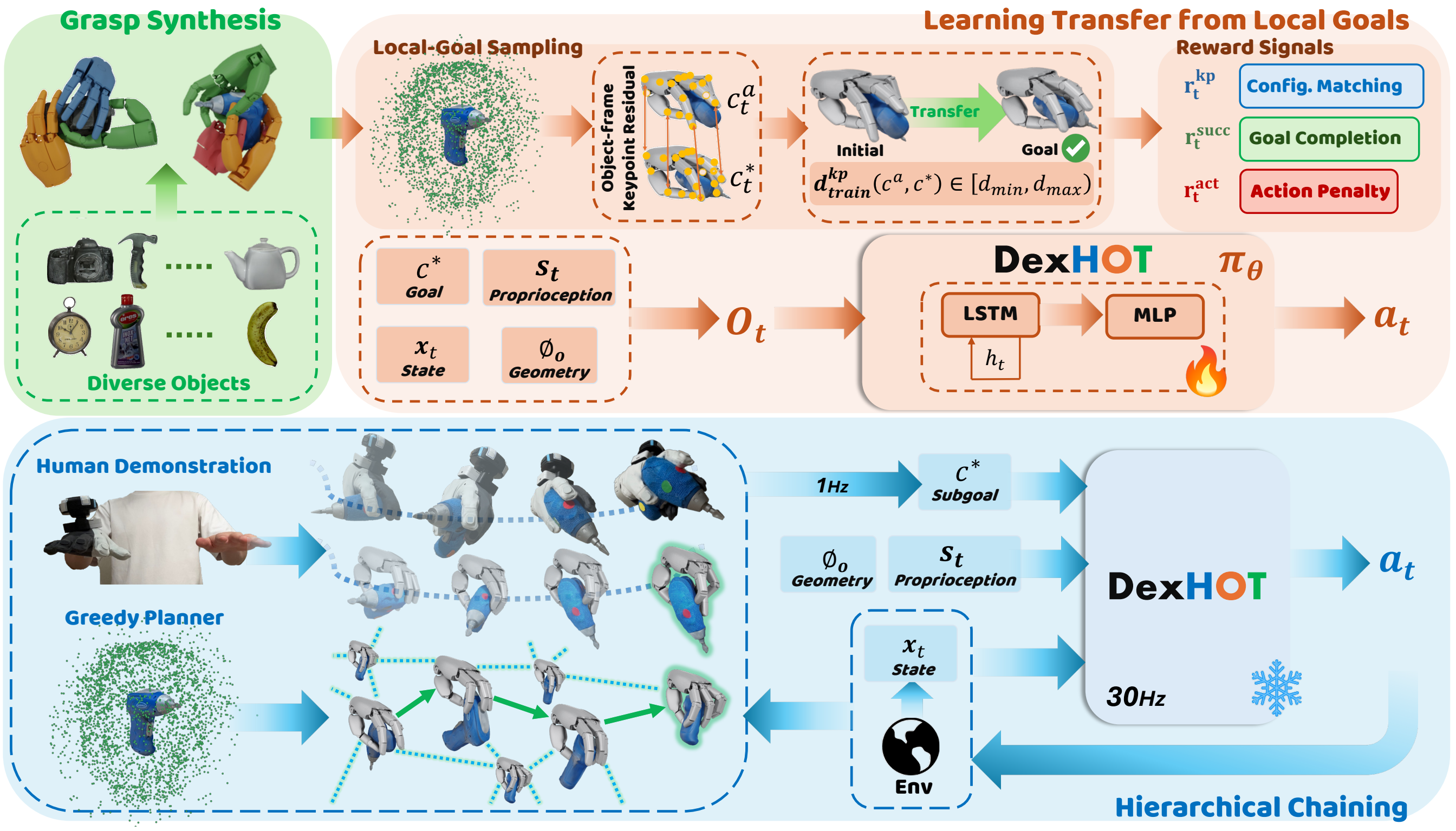
\begin{figure*}[t]
  \centering
  \input{Figures/pipeline_notation}
\caption{{\textbf{DexHOT overview}}. \emph{Top:} a recurrent policy learns local transfers from physically validated configuration pools using population-based reinforcement learning. Observations include goals, object-frame keypoint residuals, proprioception, and object geometry. The displayed reward signals encourage configuration matching, transfer completion, and action regularization. \emph{Bottom:} the frozen policy follows subgoals from greedy planning or human demonstrations through hierarchical chaining.}
  \label{fig:pipeline}
\end{figure*}

DexHOT builds on this insight in three steps (Fig.~\ref{fig:pipeline}). It synthesizes dense pools of physically feasible configurations for each object (Sec.~\ref{sec:data}), trains one policy to transfer between neighboring configurations in these pools (Sec.~\ref{sec:train}), and composes the learned transfers through hierarchical chaining to execute tasks specified by a human demonstration or a planner (Sec.~\ref{sec:deploy}). The same pools define DexHOTBench, our benchmark for configuration transfer (Sec.~\ref{sec:benchmark}).

\subsection{Feasible Configuration Synthesis}
\label{sec:data}
\label{sec:obj}
\label{sec:grasp_gen}
\label{sec:filter}
\label{sec:budget}

We construct configurations that stably hold the object, satisfy penetration constraints, and place it within the hand's manipulable region. Each retained configuration can serve as a training target or reset state.

We use 40 objects from 40 Omni6DPose categories~\citep{omni6dpose}. For each object, BODex~\citep{bodex} optimizes full-hand and fingertip grasps from surface-sampled seeds under varied object orientations, producing pre-grasp, grasp, and squeeze poses. Following the BODex validation procedure, we filter candidates in Isaac Gym~\citep{isaacgym} with our training simulation parameters and collision geometry, yielding $825{,}473$ configurations organized into per-object training pools $\mathcal{D}_o$.

\subsection{Learning Transfer from Local Goals}
\label{sec:train}

We train a shared recurrent policy network across all 40 objects using Evolutionary Policy Optimization (EPO)~\citep{epo} in Isaac Gym. Its block-conditioned policies use the active target $c_t^\star$ defined in Sec.~\ref{sec:formulation}; one block is selected for deployment.

\subsubsection{Local-Goal Sampling}
\label{sec:nearby}
Following DexterityGen~\citep{dexteritygen}, we train on nearby goals. For efficiency, we select neighbors offline: for each dataset anchor $c^a\in\mathcal{D}_o$, we precompute
\begin{equation}
\mathcal{N}(c^a)
= \bigl\{c\in\mathcal{D}_o:\ d_{\min}<d(c^a,c)\leq d_{\max}\bigr\}.
\label{eq:neighbor-pool}
\end{equation}

The tunable bounds satisfy $\tau\leq d_{\min}<d_{\max}$, excluding targets within the anchor's evaluation tolerance while controlling locality. We use $d_{\min}=20$\,mm and $d_{\max}=40$\,mm. Each episode starts from a pool configuration and samples a target from its neighborhood. After completion, the target's dataset configuration becomes the next anchor. Targets in the episode's recent history are excluded whenever unvisited neighbors remain; otherwise, revisits are allowed. The environment resets to a sampled pool configuration after a drop or a per-target timeout.

\subsubsection{Observation and Action Spaces}
\label{sec:obs}
\label{sec:action}
The goal is encoded by the active target's joint angles $\bm{q}_{h,t}^\star$ and object-frame keypoint residuals
\begin{equation}
\Delta\bm{k}_t
= \bigl[\bm{k}_j(c_t)-\bm{k}_j(c_t^\star)\bigr]_{j=1}^{K}.
\end{equation}
The $K=28$ keypoints are defined in Sec.~\ref{sec:formulation}.
Proprioception $\bm{s}_t$ includes joint angles, fingertip and end-effector poses, and the previous action. Observations also include the wrist-frame object pose and its residual to the target. The static shape descriptor $\bm{\phi}_o$ is a 64-D basis-point-set encoding~\citep{bps}; additional geometry features comprise four box-derived keypoints expressed in the current and target wrist frames~\citep{simtoolreal}, computed from the object geometry and the two configurations.

The policy outputs $\bm{a}_t\in[-1,1]^{6+n_h}$, with six unused wrist channels and $n_h$ hand channels. A separate controller supplies any required arm motion. Let $\bar{\bm{a}}_t$ denote the command after training-time action noise and delay; without these perturbations, $\bar{\bm{a}}_t=\bm{a}_t$. Following HORA~\citep{hora}, its hand channels increment the joint position targets:
\begin{equation}
\bm{q}_{h,t+1}^{\mathrm{cmd}}
= \operatorname{clip}\left(
\bm{q}_{h,t}^{\mathrm{cmd}}+\alpha_a\bar{\bm{a}}_t^h
\right).
\end{equation}
Here $n_h=22$, $\alpha_a=1/12$\,rad per control step, and clipping enforces joint limits.

\subsubsection{Reward Design}
\label{sec:reward-main}
With $d_t=d(c_t,c_t^\star)$, the base reward is
\begin{equation}
r_t^{\mathrm{base}} =
\underbrace{\exp\!\Bigl(-\frac{d_t}{2\tau_{\mathrm{tr}}}\Bigr)}_{r_t^{\mathrm{kp}}}
+\underbrace{\lambda_{\mathrm{s}}\,\mathbf{1}_t^{\mathrm{succ}}}_{r_t^{\mathrm{succ}}}
-\underbrace{\lambda_{\mathrm{a}}\lVert\bar{\bm{a}}_t\rVert_1}_{r_t^{\mathrm{act}}} .
\label{eq:transfer-reward}
\end{equation}
We set $\lambda_{\mathrm{s}}=100$ and $\lambda_{\mathrm{a}}=0.02$. The completion indicator fires once per target when $d_t\leq\tau_{\mathrm{tr}}$ holds for $H=10$ consecutive control steps, where $\tau_{\mathrm{tr}}$ is the training completion tolerance; evaluation uses $\tau=20$\,mm. The dense keypoint term encourages configuration matching, and the action penalty applies to the full 28-D command. Later in training, an additional penalty on distal-joint flexion beyond the target refines hand articulation.

\subsubsection{Population-Based Exploration}
\label{sec:algo}
In our setting, a single proximal policy optimization (PPO) policy~\citep{ppo} tends to concentrate on a limited set of configuration-transfer behaviors, restricting coverage across initial--target pairs even with many parallel environments. We therefore use EPO~\citep{epo}, which combines PPO-based updates and shared experience across 64 exploration blocks with distinct learned 32-D latent codes and shared network weights. Genetic updates preserve the designated leader and replace low-fitness follower codes through elite crossover and Gaussian mutation. In our implementation, genetic fitness is an exponential moving average of episode-level \emph{transfer throughput}: completed transfers divided by control steps. This fitness is distinct from the shaped return used for policy-gradient learning. For deployment, we select the policy with the highest empirical transfer success rate among all 64 blocks, including the leader, and freeze the shared weights and selected latent code to obtain the controller $\pi_\theta$ in \eqref{eq:policy-interface}. The same selected policy is used across all objects.

\subsubsection{Training Details}
\label{sec:dr}
The actor uses a single-layer 1024-unit long short-term memory (LSTM) network, layer normalization, and a multilayer perceptron (MLP) with exponential linear unit (ELU) activations and widths $[1024,1024,512,512]$. An asymmetric feedforward critic~\citep{asymmetric_ac} with the same MLP widths receives noise-free state with privileged joint, fingertip, and object velocities and contact information. Both networks receive the exploration block's latent code. Following prior work~\citep{simtoolreal}, we randomize object mass, scale, friction, and restitution, hand friction and link masses, and joint stiffness and damping; apply random forces and torques to the object; add noise to object-pose, joint, and fingertip observations and to actions; delay observations and actions by zero to two control steps; and ramp gravity from $0.2g$ to $g$ at the start of training. The hand is nominally palm-down; at each reset we additionally apply a random tilt of up to $30^\circ$ to the whole hand--object pair, which varies the direction of gravity relative to the hand. Training uses $524{,}288$ environments with 60\,Hz physics and 30\,Hz control.

\subsection{Hierarchical Chaining}
\label{sec:deploy}
\label{sec:sim2real}

We reuse the trained policy by supplying a sequence $\{c^{(m)}\}_{m=1}^{M}$ of configuration subgoals. When arm motion is required, the task also specifies wrist poses in the world frame $G$, giving $\{(c^{(m)},\,{}^{G}\!T_W^{(m)})\}_{m=1}^{M}$. The hand policy tracks the active subgoal $c_t^\star$ at 30\,Hz while the arm follows the corresponding wrist trajectory. On hardware, joint encoders and FoundationPose~\citep{foundationpose} reconstruct $c_t$ for feedback. Subgoals are updated at 1\,Hz, and the policy remains frozen.

\subsubsection{Subgoals from Greedy Planning}
\label{sec:planner}
For a final target $c^\star$, the planner recomputes the subgoal from $c_t$ at each update. If $d(c_t,c^\star)\leq d_{\mathrm{step}}$, it sets $c_t^\star=c^\star$. Otherwise, it selects
\begin{equation}
c' \in
\underset{\substack{c\in\mathcal{D}_o\\
\tau\leq d(c_t,c)\leq d_{\mathrm{step}}\\
d(c,c^\star)<d(c_t,c^\star)}}
{\arg\min}\; d(c,c^\star),
\label{eq:greedy-subgoal}
\end{equation}
and sets $c_t^\star=c'$. If the candidate set is empty, the planner falls back to a larger search range. We use a nominal $d_{\mathrm{step}}=50$\,mm in simulation and on the real robot. Selected configurations are validated endpoints, but the greedy rule does not guarantee the dynamic feasibility of each transition.

\subsubsection{Subgoals from Human Demonstrations}
A single human demonstration supplies the relations required by a task. Following the perception pipeline of SimToolReal~\citep{simtoolreal}, we obtain an object mesh using SAM 3D~\citep{sam3d} and use FoundationPose~\citep{foundationpose} with the mesh and RGB-D video to estimate the object's 6D pose trajectory. Manus hand motion is retargeted to the robot and refined with contact and penetration losses. We first follow the retargeted trajectory up to a designated pre-grasp frame, then switch hand control to the frozen policy for pickup and subsequent transfers. Thereafter, the aligned trajectories provide configuration subgoals in temporal order at 1\,Hz, while a separate arm controller continues to track the demonstration-derived wrist trajectory. Thus, the demonstration specifies the approach, wrist motion, and desired relations; the policy generates closed-loop hand actions from pickup onward.

\subsection{DexHOTBench}
\label{sec:benchmark}

\begin{figure}[t]
\centering
\begingroup
\setbox0=\hbox{\includegraphics{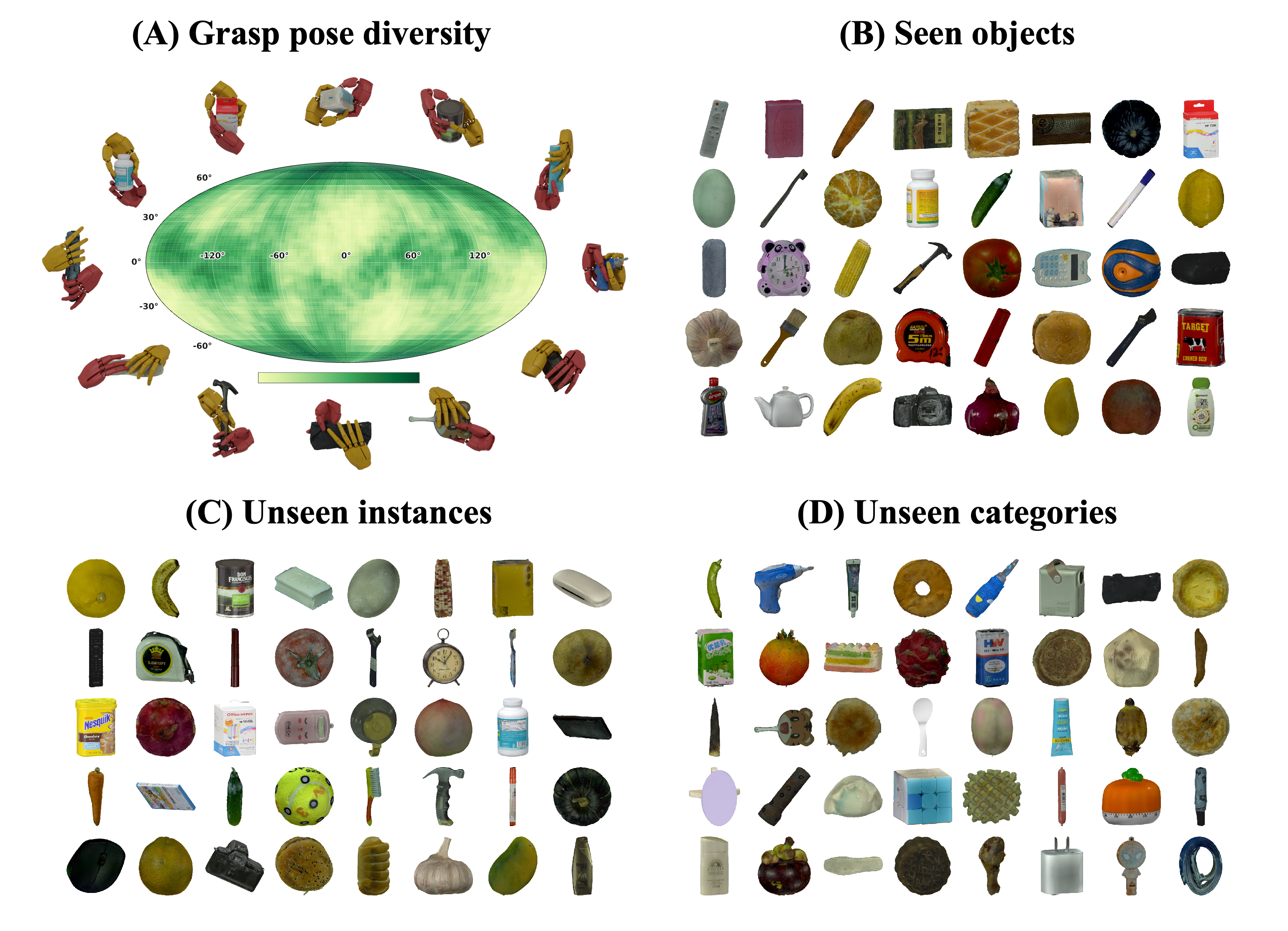}}%
\resizebox{\linewidth}{!}{%
  \rlap{\copy0}%
  \pdfliteral{q 1 1 1 rg}%
  \rlap{\kern-1bp\raisebox{-1bp}[0pt][0pt]{\rule{\dimexpr\wd0+2bp\relax}{5bp}}}%
  \rlap{\kern-1bp\raisebox{\dimexpr\ht0-4bp\relax}[0pt][0pt]{\rule{\dimexpr\wd0+2bp\relax}{5bp}}}%
  \rlap{\kern-1bp\raisebox{-1bp}[0pt][0pt]{\rule{5bp}{\dimexpr\ht0+2bp\relax}}}%
  \rlap{\kern\dimexpr\wd0-4bp\relax\raisebox{-1bp}[0pt][0pt]{\rule{5bp}{\dimexpr\ht0+2bp\relax}}}%
  \pdfliteral{Q}%
  \phantom{\copy0}}%
\endgroup
\caption{\textbf{DexHOTBench.} \textbf{(A)} Object $z$-axis directions relative to the hand across all configuration pools, shown on a Mollweide map colored by density percentile, from pale yellow-green (lower density) to dark green (higher density). Rotation about the object's $z$-axis is omitted. \textbf{(B--D)} Seen objects, unseen instances of training categories, and objects from unseen categories, respectively; $40$ objects per split.}
\label{fig:benchmark}
\end{figure}

Following the object-category taxonomy of Omni6DPose~\citep{omni6dpose} and the object-novelty split used by UniDexGrasp~\citep{unidexgrasp}, \textbf{DexHOTBench} (Fig.~\ref{fig:benchmark}) evaluates configuration transfer on 120 objects: 40 training objects, 40 held-out objects from the training categories (one per category), and 40 held-out objects from 40 new categories. The 80 held-out objects do not appear in policy training. Each held-out object has its own configuration pool, produced by the same synthesis and validation procedure (Sec.~\ref{sec:data}). The 120 pools contain $1{,}184{,}058$ configurations, including $358{,}585$ for held-out objects. We retain objects with convex-hull bounding-sphere diameter at most 300\,mm and minimum convex-hull extent at least 10\,mm. Larger meshes rarely passed the grasp filter and were unstable in simulation, whereas thinner meshes caused interpenetration and increased contact-computation cost.

Pairs are grouped by $d(c_0,c^\star)$ into \textbf{Neighbor} $(20,40]$\,mm, matching the training sampling band used here, and \textbf{Any2Any} close $(20,70]$\,mm and far $(70,\infty)$\,mm. We independently sample and fix 1000 pairs per object and range; Neighbor and close ranges overlap, and seen-object endpoints come from the training pools. Evaluation uses the planner of Sec.~\ref{sec:planner}, which commands targets within its step bound directly.

Each pair is evaluated once with the same policy weights, latent code, and nominal physics, using the success criterion of Sec.~\ref{sec:formulation} and a budget of 20 target updates. Timer-driven replanning occurs at 1\,Hz, while target completion can trigger an earlier update. Success rates are averaged across objects. We will release the objects, configuration pools, evaluation pairs, and scoring script.

%% file: Figures/pipeline_notation.tex
\includegraphics[width=\textwidth]{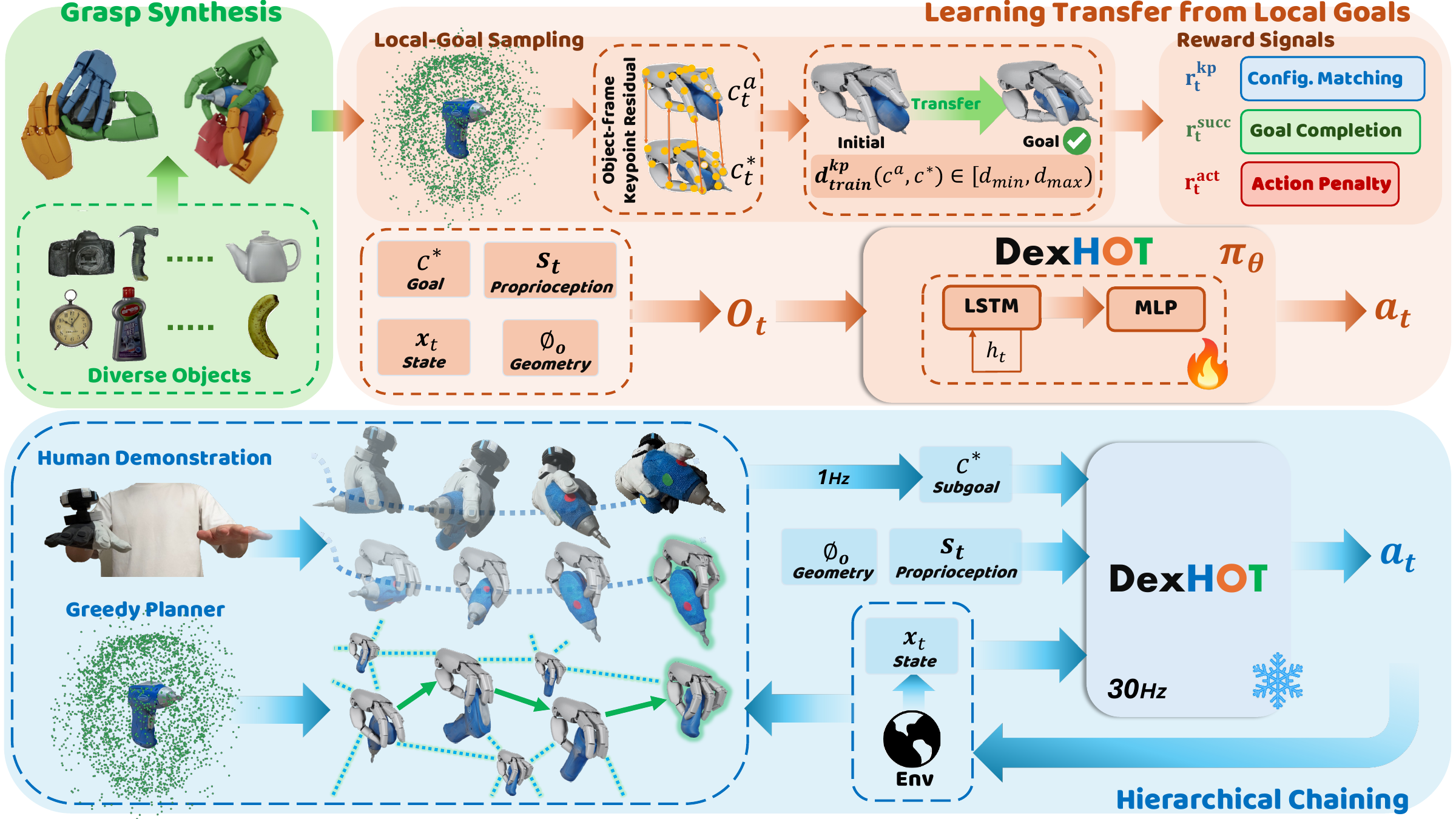}

%% file: Texs/experiments.tex
\section{Experiments}
\label{sec:experiments}

We evaluate DexHOT in simulation and on real hardware to assess Any2Any hand--object configuration transfer across objects and its extension to long-horizon manipulation. Our experiments address four questions: \textbf{Q1:} How well does DexHOT generalize across transfer distances and objects? \textbf{Q2:} Which training choices contribute to configuration-transfer success? \textbf{Q3:} Can DexHOT perform real-world configuration transfer without fine-tuning? \textbf{Q4:} Can the same policy execute long-horizon tasks from human demonstrations?
\subsection{Experimental Setup}
\label{sec:exp-setup}

\subsubsection{Simulation Protocol}
We evaluate all 40 training and 80 held-out DexHOTBench objects (Sec.~\ref{sec:benchmark}) with one checkpoint and a fixed latent code, without object-specific policy weights. Each held-out object uses its own offline-generated configuration pool for evaluation endpoints and planner candidates, while the policy weights, latent code, and planner parameters remain fixed across objects. All configuration-transfer evaluations use the greedy planner of Sec.~\ref{sec:planner}, with $d_{\mathrm{step}}=50$\,mm and 1\,Hz replanning.

\subsubsection{Real-Robot Setup}
We use the same 22-degree-of-freedom (DoF) Sharpa Wave hand as in simulation, mounted on a 7-DoF Franka Emika Panda arm. Joint encoders and FoundationPose~\citep{foundationpose}, using a third-person Intel RealSense D435i RGB-D camera, provide the current configuration $c_t$ in \eqref{eq:configuration}. The policy runs at 30\,Hz without fine-tuning. Configuration-transfer trials (Sec.~\ref{sec:exp-real-anytoany}) fix the arm in a canonical palm-down pose to evaluate finger-driven motion. In demonstration-guided tasks (Sec.~\ref{sec:exp-real-hoi}), the arm also tracks the demonstrated wrist trajectory at 30\,Hz while the policy controls the hand--object configuration.

\subsection{Transfer Range and Object Generalization}
\label{sec:exp-sim-results}

\subsubsection{Transfer Beyond the Training Band}
Table~\ref{tab:sim-main} addresses \textbf{Q1}. On the 40 seen objects, DexHOT achieves 90.9\% success on Neighbor transfers, 85.5\% on close Any2Any transfers, and 75.6\% on far transfers. The close range extends to 70\,mm, beyond the local-goal sampling band. Far-transfer performance evaluates the learned controller with hierarchical chaining. These results show that training on local goals supports transfer across the evaluated distance range.

\begin{table}[!t]
\centering
\caption{Simulation success rates by initial configuration distance (mm). Counts in parentheses denote the number of evaluated objects. Seen objects are objects from the training set; unseen instances are held-out objects from training categories; unseen categories contain one object from each of $40$ held-out categories. Avg.\ denotes the mean of the close and far Any2Any success rates.}
\label{tab:sim-main}
\input{Tables/sim_main}
\end{table}

\subsubsection{Generalization to Unseen Objects}
Averaged over close and far transfers, success decreases from 80.5\% on seen objects to 74.4\% on unseen objects of seen categories and 64.4\% on unseen categories. Performance also decreases with transfer distance in every split: on unseen objects of seen categories and on unseen categories, Neighbor success is 88.3\% and 81.5\%, respectively, while far-transfer success is 67.3\% and 56.7\%. The most challenging condition combines an unseen category with a large configuration change.

\subsection{Training Ablations}
\label{sec:exp-ablations}

\subsubsection{Ablation Design}
To address \textbf{Q2}, we vary the exploration algorithm, goal-sampling distribution, and configuration used to anchor goal sampling. Table~\ref{tab:ablations} evaluates all variants on the same 40 seen objects under the Neighbor and Any2Any protocols. All variants use the same greedy planner and evaluation protocol. The Any2Any columns isolate training choices under a fixed greedy planner; they are not a direct-to-final versus chaining planner ablation.

\begin{table}[!htb]
\centering
\caption{Training ablations on $40$ seen objects using the same evaluation protocol as Table~\ref{tab:sim-main}. Bold marks DexHOT.}
\label{tab:ablations}
\input{Tables/ablations}
\end{table}

\subsubsection{Population-Based Exploration}
The \textsc{w/o EPO} variant replaces EPO with a single PPO agent~\citep{ppo} at matched environment counts. Even its best training checkpoint reaches only 11.5\% success on Neighbor transfers and 2.5\% on close transfers, and it completes no far transfer, supporting the importance of population-based exploration in this setting.

\subsubsection{Local-Goal Sampling}
The \textsc{w/o local-goal sampling} variant samples targets uniformly from the full configuration pool. It performs worse in every distance range, including far transfers, despite exposure to distant targets. This supports local-goal sampling as an effective training strategy across the evaluated transfer distances.

\subsubsection{Dataset and Online Anchors}
The \textsc{w/ online anchor} variant forms neighborhoods around the current configuration $c_t$ instead of the dataset anchor $c^a$ (Sec.~\ref{sec:nearby}). Neighbor success remains high, but far-transfer success drops from 75.6\% to 52.8\%, favoring dataset anchoring for distant transfers. Dataset anchors also avoid online neighborhood searches.

\subsection{Real-World Configuration Transfer}
\label{sec:exp-real-anytoany}

\begin{figure*}[!t]
\centering
\includegraphics[width=\textwidth]{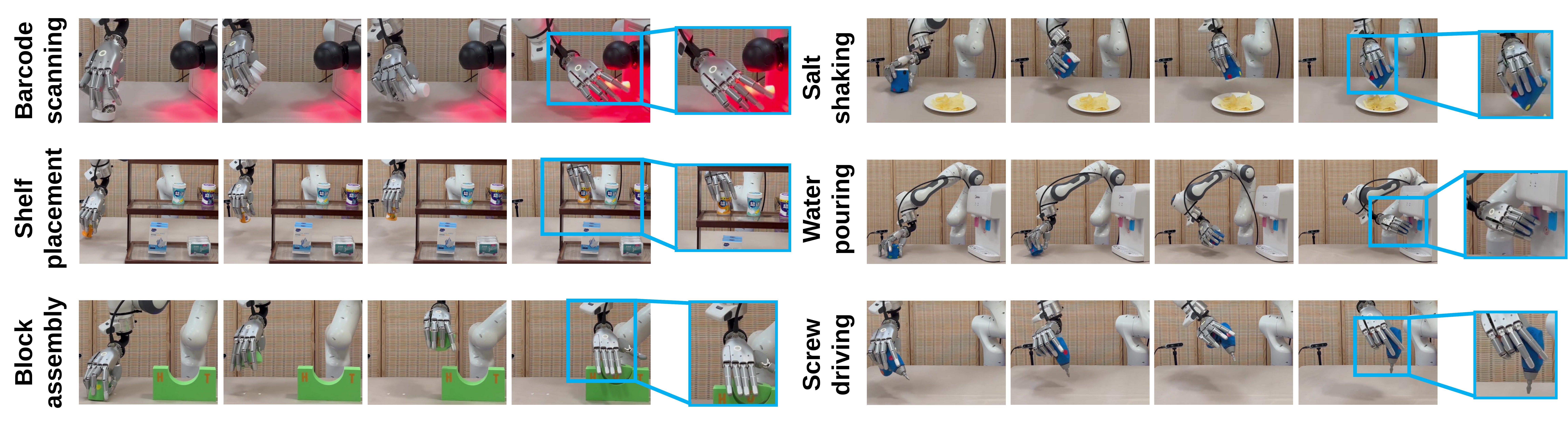}
\caption{Demonstration-guided real-robot tasks. Each row within a panel shows one task, with frames ordered left to right; insets enlarge the final hand--object configuration.}
\label{fig:task2}
\end{figure*}

\subsubsection{Evaluation Protocol}
For \textbf{Q3}, we test five seen and five unseen objects. Each object's pool supplies 20 random initial--target pairs, one attempt per pair: 200 trials spanning small and large changes. Planner settings match simulation.

\subsubsection{Evaluation Metrics}
Each trial first attempts to reproduce the initial configuration and then transfer to $c^\star$. Following Sec.~\ref{sec:formulation}, success requires $d(c_t,c^\star)\leq\tau$ for $H=10$ consecutive control steps without dropping the object, with $\tau=20$\,mm. Let $d_{\mathrm{init}}$ denote the configuration distance to $c^\star$ at the start of the transfer. We define $d_{\mathrm{hold}}$ as the minimum, over all complete $H$-step windows during the transfer, of the maximum final-target distance $d(c_t,c^\star)$ within each window. For $d_{\mathrm{init}}>\tau$, transfer progress is
\begin{equation}
P_{\mathrm{transfer}} = \operatorname{clip}_{[0,1]}\!\left(
\frac{d_{\mathrm{init}}-d_{\mathrm{hold}}}{d_{\mathrm{init}}-\tau}
\right).
\label{eq:transfer-progress}
\end{equation}
This measures the fraction of the required distance reduction achieved and sustained for $H$ consecutive control steps. All pairs satisfy $d_{\mathrm{init}}>\tau$. Trials that fail to reproduce the initial configuration count as failures with zero transfer progress.

\subsubsection{Transfer Performance}
DexHOT achieves 61.5\% overall success and average transfer progress of 0.735 across the 200 real-robot transfers (Table~\ref{tab:real-anytoany}). The seen and unseen sets reach similar success rates of 63\% and 60\%, supporting transfer to these evaluated unseen objects. The tennis ball and medicine bottle achieve the highest success rates, whereas the screwdriver and hammer are the most challenging. Failures on these elongated tools include wedging between fingers and lateral swinging during final alignment.

\begin{table}[!htb]
\centering
\caption{Real-world configuration transfer over 20 trials per object. SR: success rate; Prog.: mean transfer progress \eqref{eq:transfer-progress}. Initialization failures count as failures with zero progress.}
\label{tab:real-anytoany}
\input{Tables/real_anytoany}
\end{table}

\subsubsection{Failure Analysis}
We observed four failure modes:
(i) \emph{Slippage without recovery.} On low-friction objects, slips or rotations can develop too quickly for closed-loop adjustment to recover. Higher-rate feedback or tactile sensing may help.
(ii) \emph{Initialization failure.} Some configurations established by pre-grasp interpolation and squeezing (Sec.~\ref{sec:data}) are marginally stable on hardware and cannot be reproduced accurately enough to begin transfer.
(iii) \emph{Stalled reorientation.} Progress stops before reaching the target, possibly due to friction mismatch, lateral Sharpa-joint motion, or discrepancies between simulated and physical collision geometry.
(iv) \emph{Object-tracking loss.} FoundationPose occasionally loses tracking, leaving no valid pose estimate. More frequent failures on complex geometries motivate richer shape representations and improved dynamics modeling and sensing.

\subsection{Long-Horizon Tasks from Human Demonstrations}
\label{sec:exp-real-hoi}

\subsubsection{Evaluation Protocol}
For \textbf{Q4}, hierarchical chaining (Sec.~\ref{sec:deploy}) executes six tasks with the same controller. Each task uses one retargeted demonstration and five trials from a fixed initial layout. We first follow the retargeted demonstration trajectory up to its designated pre-grasp frame, then switch hand control to the frozen policy for pickup and subsequent configuration transfers. From this frame onward, DexHOT tracks configuration subgoals in closed loop at 30\,Hz, with subgoal updates at 1\,Hz, while the arm continues to track the demonstration-derived wrist trajectory.

\subsubsection{Task Suite}
Fig.~\ref{fig:task2} illustrates the tasks. Barcode scanning and shelf placement rotate a bottle by $180^\circ$, with shelf placement requiring a pinch-to-power grasp transition. Block assembly reorients a half-round block to align it with a slot. Salt shaking turns the shaker dispensing-face down and back up. Water pouring turns an inverted cup upright to receive water, and screw driving reorients the drill bit downward.

\subsubsection{Evaluation Metrics}
We evaluate three equally weighted task stages:
(i) picking up the object;
(ii) satisfying the task-specific object-level constraint, such as directing the barcode toward the scanner, the cup opening upward, or the drill bit downward; and
(iii) attaining the required hand--object configuration, including the task-specific hand arrangement.
Task success requires completing all three stages. Stage progress is the fraction of completed stages, averaged over the five trials for each task, and is distinct from the geometric transfer progress in \eqref{eq:transfer-progress}. Stage-wise success counts use all trials as the denominator.

\subsubsection{Results and Baseline Comparisons}
\begin{table}[!tb]
\centering
\caption{Demonstration-guided task execution from fixed initial layouts. SR: successes out of five trials per task; the Average row reports success-rate fractions. Prog.: mean fraction of task stages completed. \textbf{Bold} marks DexHOT.}
\label{tab:real-hoi}
\input{Tables/real_hoi}
\end{table}
Table~\ref{tab:real-hoi} reports results under this criterion. DexHOT completes all three stages in 27/30 trials (90\%), including all trials of shelf placement, block assembly, salt shaking, and water pouring.

Kinematic retargeting uses a tuned Manus-to-Sharpa mapping, followed by trajectory optimization to reduce self-penetration and hand--object penetration while promoting temporal smoothness. The resulting joint trajectory is replayed without closed-loop hand--object correction. This baseline succeeds at pickup in 24/30 trials but fails to follow the subsequent hand--object configuration sequence, with slipping and dropping during reorientation. Neither stage (ii) nor stage (iii) is completed in any trial (0/30 each).

We evaluate a per-object adaptation of SimToolReal~\citep{simtoolreal}, retaining its joint arm--hand control interface. In contrast, DexHOT controls the hand while a separate arm controller tracks the demonstration-derived wrist trajectory. Both setups are evaluated using the same three-stage task criterion. SimToolReal completes pickup in 21/30 trials and satisfies the object-level constraint in 9/30 trials. Only 2/30 trials, both in block assembly, also attain the required hand--object configuration and satisfy the full three-stage criterion, despite that configuration not being an explicit policy goal.

The gap between object-level and configuration-level completion illustrates that attaining an object-pose goal does not ensure the specified hand arrangement. DexHOT achieves higher task success under the shared three-stage evaluation criterion.

%% file: Tables/sim_main.tex
\begingroup
\footnotesize
\renewcommand{\arraystretch}{1.1}
\setlength{\tabcolsep}{1.5pt}
\begin{tabularx}{\linewidth}{@{}l*{4}{>{\centering\arraybackslash}X}@{}}
\toprule
\multirow{2}{*}{\textbf{Object set}}
 & \textbf{Neighbor}
 & \multicolumn{2}{c}{\textbf{Any2Any}}
 & \multirow{2}{*}{\textbf{Avg.}} \\
\cmidrule(lr){2-2}\cmidrule(lr){3-4}
 & $(20,40]$ & $(20,70]$ & $(70,\infty)$ & \\
\midrule
Seen objects (40)       & 0.909 & 0.855 & 0.756 & 0.805 \\
Unseen instances (40)   & 0.883 & 0.814 & 0.673 & 0.744 \\
Unseen categories (40)  & 0.815 & 0.722 & 0.567 & 0.644 \\
\bottomrule
\end{tabularx}
\endgroup

%% file: Tables/ablations.tex
\begingroup
\footnotesize
\renewcommand{\arraystretch}{1.1}
\setlength{\tabcolsep}{2.5pt}
\begin{tabularx}{\linewidth}{@{}l*{4}{>{\centering\arraybackslash}X}@{}}
\toprule
 & \textbf{Neighbor} & \multicolumn{2}{c}{\textbf{Any2Any}} & \\
\cmidrule(lr){2-2}\cmidrule(lr){3-4}
\textbf{Method} & (20,40] & (20,70] & (70,$\infty$) & \textbf{Avg.} \\
\midrule
DexHOT (Ours) & \textbf{0.909} & \textbf{0.855} & \textbf{0.756} & \textbf{0.805} \\
w/o EPO           & 0.115 & 0.025 & 0.000 & 0.013 \\
w/o local-goal sampling & 0.841 & 0.752 & 0.612 & 0.682 \\
w/ online anchor & 0.860 & 0.751 & 0.528 & 0.640 \\
\bottomrule
\end{tabularx}
\endgroup

%% file: Tables/real_anytoany.tex
\begingroup
\footnotesize
\renewcommand{\arraystretch}{1.1}
\setlength{\tabcolsep}{2pt}
\begin{tabularx}{\linewidth}{@{}>{\raggedright\arraybackslash}Xcc@{\hspace{3pt}}>{\raggedright\arraybackslash}Xcc@{}}
\toprule
\multicolumn{3}{c}{\textbf{Seen objects}} & \multicolumn{3}{c}{\textbf{Unseen objects}} \\
\cmidrule(lr){1-3}\cmidrule(lr){4-6}
\textbf{Object} & \textbf{SR} & \textbf{Prog.} & \textbf{Object} & \textbf{SR} & \textbf{Prog.} \\
\midrule
\makebox[6mm][c]{\raisebox{-0.3\height}{\includegraphics[width=6mm,height=5.5mm,keepaspectratio]{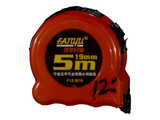}}}\hspace{2pt}\mbox{Tape measure} & 0.80 & 0.918 & \makebox[6mm][c]{\raisebox{-0.3\height}{\includegraphics[width=6mm,height=5.5mm,keepaspectratio]{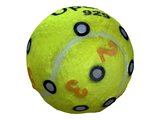}}}\hspace{2pt}\mbox{Tennis ball} & 0.95 & 0.950 \\
\makebox[6mm][c]{\raisebox{-0.3\height}{\includegraphics[width=6mm,height=5.5mm,keepaspectratio]{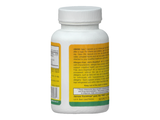}}}\hspace{2pt}\mbox{Medicine bottle} & 0.90 & 0.918 & \makebox[6mm][c]{\raisebox{-0.3\height}{\includegraphics[width=6mm,height=5.5mm,keepaspectratio]{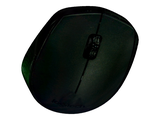}}}\hspace{2pt}\mbox{Mouse} & 0.70 & 0.760 \\
\makebox[6mm][c]{\raisebox{-0.3\height}{\includegraphics[width=6mm,height=5.5mm,keepaspectratio]{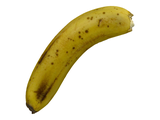}}}\hspace{2pt}\mbox{Banana} & 0.80 & 0.837 & \makebox[6mm][c]{\raisebox{-0.3\height}{\includegraphics[width=6mm,height=5.5mm,keepaspectratio]{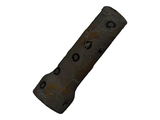}}}\hspace{2pt}\mbox{Flashlight} & 0.70 & 0.831 \\
\makebox[6mm][c]{\raisebox{-0.3\height}{\includegraphics[width=6mm,height=5.5mm,keepaspectratio]{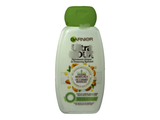}}}\hspace{2pt}\mbox{Shampoo bottle} & 0.45 & 0.677 & \makebox[6mm][c]{\raisebox{-0.3\height}{\includegraphics[width=6mm,height=5.5mm,keepaspectratio]{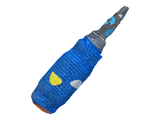}}}\hspace{2pt}\mbox{Screwdriver} & 0.20 & 0.405 \\
\makebox[6mm][c]{\raisebox{-0.3\height}{\includegraphics[width=6mm,height=5.5mm,keepaspectratio]{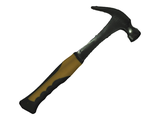}}}\hspace{2pt}\mbox{Hammer} & 0.20 & 0.377 & \makebox[6mm][c]{\raisebox{-0.3\height}{\includegraphics[width=6mm,height=5.5mm,keepaspectratio]{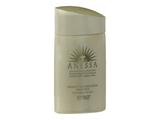}}}\hspace{2pt}\mbox{Facial cream} & 0.45 & 0.675 \\
\midrule
Seen average & 0.63 & 0.745 & Unseen average & 0.60 & 0.724 \\
\bottomrule
\end{tabularx}
\endgroup

%% file: Tables/real_hoi.tex
\begingroup
\footnotesize
\renewcommand{\arraystretch}{1.1}
\setlength{\tabcolsep}{3pt}
\begin{tabularx}{\linewidth}{@{}l*{6}{>{\centering\arraybackslash}X}@{}}
\toprule
 & \multicolumn{2}{c}{\textbf{Kin.~Retarget}} & \multicolumn{2}{c}{\textbf{SimToolReal}} & \multicolumn{2}{c}{\textbf{DexHOT (Ours)}} \\
\cmidrule(lr){2-3}\cmidrule(lr){4-5}\cmidrule(lr){6-7}
\textbf{Task} & SR & Prog. & SR & Prog. & SR & Prog. \\
\midrule
Barcode scanning & 0/5 & 0.33 & 0/5 & 0.53 & \textbf{4/5} & \textbf{0.93} \\
Shelf placement  & 0/5 & 0.33 & 0/5 & 0.33 & \textbf{5/5} & \textbf{1.00} \\
Block assembly   & 0/5 & 0.27 & 2/5 & 0.73 & \textbf{5/5} & \textbf{1.00} \\
Salt shaking     & 0/5 & 0.33 & 0/5 & 0.33 & \textbf{5/5} & \textbf{1.00} \\
Water pouring    & 0/5 & 0.00 & 0/5 & 0.00 & \textbf{5/5} & \textbf{1.00} \\
Screw driving    & 0/5 & 0.33 & 0/5 & 0.20 & \textbf{3/5} & \textbf{0.80} \\
\midrule
Average & 0.00 & 0.27 & 0.07 & 0.36 & \textbf{0.90} & \textbf{0.96} \\
\bottomrule
\end{tabularx}
\endgroup

%% file: Texs/conclusion.tex
\section{Conclusion}
\label{sec:conclusion}

DexHOT learns reusable hand--object configuration transfers from pairs of nearby, synthetically generated configurations represented by hand keypoints in the object frame. Hierarchical chaining with greedy planning reaches more distant goals; demonstration-derived subgoals support tasks without policy fine-tuning. Averaged over close and far transfers, simulation success is 80.5\% on seen objects and 69.4\% on unseen objects; real-robot success is 61.5\% over 200 transfers. Together with the six demonstration-guided tasks, these results support hand--object configurations as a practical interface for dexterous behavior.

%% file: main.bbl
\begin{thebibliography}{10}
\providecommand{\url}[1]{#1}
\csname url@rmstyle\endcsname
\providecommand{\newblock}{\relax}
\providecommand{\bibinfo}[2]{#2}
\providecommand\BIBentrySTDinterwordspacing{\spaceskip=0pt\relax}
\providecommand\BIBentryALTinterwordstretchfactor{4}
\providecommand\BIBentryALTinterwordspacing{\spaceskip=\fontdimen2\font plus
\BIBentryALTinterwordstretchfactor\fontdimen3\font minus
  \fontdimen4\font\relax}
\providecommand\BIBforeignlanguage[2]{{%
\expandafter\ifx\csname l@#1\endcsname\relax
\typeout{** WARNING: IEEEtran.bst: No hyphenation pattern has been}%
\typeout{** loaded for the language `#1'. Using the pattern for}%
\typeout{** the default language instead.}%
\else
\language=\csname l@#1\endcsname
\fi
#2}}

\bibitem{clutterdexgrasp}
Z.~Chen, Q.~Yan, Y.~Chen, T.~Wu, J.~Zhang, Z.~Ding, J.~Li, Y.~Yang, and
  H.~Dong, ``{ClutterDexGrasp}: A sim-to-real system for general dexterous
  grasping in cluttered scenes,'' in \emph{Conference on Robot Learning
  (CoRL)}, 2025.

\bibitem{graspgf}
T.~Wu, M.~Wu, J.~Zhang, Y.~Gan, and H.~Dong, ``Learning score-based grasping
  primitive for human-assisting dexterous grasping,'' in \emph{Advances in
  Neural Information Processing Systems (NeurIPS)}, vol.~36, 2023, pp.
  22\,132--22\,150.

\bibitem{cadgrasp}
J.~Zhang, Z.~Ma, T.~Wu, Z.~Chen, and H.~Dong, ``{CADGrasp}: Learning contact
  and collision aware general dexterous grasping in cluttered scenes,'' in
  \emph{Advances in Neural Information Processing Systems (NeurIPS)}, vol.~38,
  2025, pp. 33\,764--33\,787.

\bibitem{openai_dactyl}
{OpenAI}, M.~Andrychowicz, B.~Baker, M.~Chociej, R.~Jozefowicz, B.~McGrew,
  J.~Pachocki, A.~Petron, M.~Plappert, G.~Powell, A.~Ray, J.~Schneider,
  S.~Sidor, J.~Tobin, P.~Welinder, L.~Weng, and W.~Zaremba, ``Learning
  dexterous in-hand manipulation,'' \emph{arXiv preprint arXiv:1808.00177},
  2018.

\bibitem{openai_rubiks}
I.~Akkaya, M.~Andrychowicz, M.~Chociej, M.~Litwin, B.~McGrew, A.~Petron,
  A.~Paino, M.~Plappert, G.~Powell, R.~Ribas, J.~Schneider, N.~Tezak,
  J.~Tworek, P.~Welinder, L.~Weng, Q.~Yuan, W.~Zaremba, and L.~Zhang, ``Solving
  {Rubik}'s cube with a robot hand,'' \emph{arXiv preprint arXiv:1910.07113},
  2019.

\bibitem{visual_dexterity}
T.~Chen, M.~Tippur, S.~Wu, V.~Kumar, E.~H. Adelson, and P.~Agrawal, ``Visual
  dexterity: In-hand reorientation of novel and complex object shapes,''
  \emph{Science Robotics}, vol.~8, no.~84, p. eadc9244, 2023.

\bibitem{hora}
H.~Qi, A.~Kumar, R.~Calandra, Y.~Ma, and J.~Malik, ``In-hand object rotation
  via rapid motor adaptation,'' in \emph{Conference on Robot Learning (CoRL)},
  2022.

\bibitem{rotateit}
H.~Qi, B.~Yi, S.~Suresh, M.~Lambeta, Y.~Ma, R.~Calandra, and J.~Malik,
  ``General in-hand object rotation with vision and touch,'' in
  \emph{Conference on Robot Learning (CoRL)}, 2023.

\bibitem{drop}
A.~H. Li, P.~Culbertson, V.~Kurtz, and A.~D. Ames, ``{DROP}: Dexterous
  reorientation via online planning,'' \emph{arXiv preprint arXiv:2409.14562},
  2024.

\bibitem{simple2complex}
H.~Qi, B.~Yi, M.~Lambeta, Y.~Ma, R.~Calandra, and J.~Malik, ``From simple to
  complex skills: The case of in-hand object reorientation,'' \emph{arXiv
  preprint arXiv:2501.05439}, 2025.

\bibitem{simtoolreal}
K.~Kedia, T.~G.~W. Lum, J.~Bohg, and K.~Liu, ``{SimToolReal}: An object-centric
  policy for zero-shot dexterous tool manipulation,'' in \emph{Proceedings of
  Robotics: Science and Systems}, Sydney, Australia, July 2026.

\bibitem{dexteritygen}
Z.-H. Yin, C.~Wang, L.~Pineda, F.~Hogan, K.~Bodduluri, A.~Sharma, P.~Lancaster,
  I.~Prasad, M.~Kalakrishnan, J.~Malik, M.~Lambeta, T.~Wu, P.~Abbeel, and
  M.~Mukadam, ``{DexterityGen}: Foundation controller for unprecedented
  dexterity,'' in \emph{Proceedings of Robotics: Science and Systems}, 2025.

\bibitem{wu2025canonical3dtactile}
T.~Wu, J.~Li, J.~Zhang, M.~Wu, and H.~Dong, ``Canonical representation and
  force-based pretraining of 3d tactile for dexterous visuo-tactile policy
  learning,'' in \emph{IEEE International Conference on Robotics and Automation
  (ICRA)}, 2025.

\bibitem{li2025adaptivevisuotactile}
J.~Li, T.~Wu, J.~Zhang, Z.~Chen, H.~Jin, M.~Wu, Y.~Shen, Y.~Yang, and H.~Dong,
  ``Adaptive visuo-tactile fusion with predictive force attention for dexterous
  manipulation,'' in \emph{IEEE/RSJ International Conference on Intelligent
  Robots and Systems (IROS)}, 2025.

\bibitem{dextreme}
A.~Handa, A.~Allshire, V.~Makoviychuk, A.~Petrenko, R.~Singh, J.~Liu,
  D.~Makoviichuk, K.~Van~Wyk, A.~Zhurkevich, B.~Sundaralingam, J.-F. Lafleche,
  D.~Fox, Y.~Narang, and G.~State, ``{DeXtreme}: Transfer of agile in-hand
  manipulation from simulation to reality,'' \emph{arXiv preprint
  arXiv:2210.13702}, 2022.

\bibitem{anyrotate}
M.~Yang, C.~Lu, A.~Church, Y.~Lin, C.~Ford, H.~Li, E.~Psomopoulou, D.~A.
  Barton, and N.~F. Lepora, ``{AnyRotate}: Gravity-invariant in-hand object
  rotation with sim-to-real touch,'' in \emph{Conference on Robot Learning
  (CoRL)}, 2024.

\bibitem{touch_dexterity}
Z.-H. Yin, B.~Huang, Y.~Qin, Q.~Chen, and X.~Wang, ``Rotating without seeing:
  Towards in-hand dexterity through touch,'' in \emph{Robotics: Science and
  Systems (RSS)}, 2023.

\bibitem{dexndm}
X.~Liu, H.~Wang, and L.~Yi, ``{DexNDM}: Closing the reality gap for dexterous
  in-hand rotation via joint-wise neural dynamics model,'' \emph{arXiv preprint
  arXiv:2510.08556}, 2025.

\bibitem{dexpoint}
Y.~Qin, B.~Huang, Z.-H. Yin, H.~Su, and X.~Wang, ``{DexPoint}: Generalizable
  point cloud reinforcement learning for sim-to-real dexterous manipulation,''
  \emph{arXiv preprint arXiv:2211.09423}, 2022.

\bibitem{opendexgrasp}
J.~Zhang, J.~Wang, T.~Wang, Z.~Chen, A.~Bolton, Y.~Peng, and H.~Dong,
  ``{OpenDexGrasp}: Open-vocabulary task-oriented dexterous grasping,'' in
  \emph{Conference on Robot Learning (CoRL)}, 2026, to appear.

\bibitem{long2025checkmanual}
Y.~Long, J.~Zhang, M.~Pan, T.~Wu, T.~Kim, and H.~Dong, ``{CheckManual}: A new
  challenge and benchmark for manual-based appliance manipulation,'' in
  \emph{Proceedings of the IEEE/CVF Conference on Computer Vision and Pattern
  Recognition (CVPR)}, 2025.

\bibitem{zeng2024lvdiffusor}
Y.~Zeng, M.~Wu, L.~Yang, J.~Zhang, H.~Ding, H.~Cheng, and H.~Dong,
  ``{LVDiffusor}: Distilling functional rearrangement priors from large models
  into diffusor,'' \emph{IEEE Robotics and Automation Letters}, vol.~9, no.~10,
  pp. 8258--8265, 2024.

\bibitem{long2026scalingmanual}
Y.~Long, L.~Kang, Z.~Yu, Y.~Gao, B.~Cheng, J.~Zhang, X.~Li, H.~Yang, D.~Li,
  H.~Shen, and H.~Dong, ``Scaling manual-grounded appliance manipulation with
  data synthesis and unified planning,'' \emph{arXiv preprint
  arXiv:2608.15863}, 2026.

\bibitem{han_gaiting}
L.~Han and J.~C. Trinkle, ``Dextrous manipulation by rolling and finger
  gaiting,'' in \emph{IEEE International Conference on Robotics and Automation
  (ICRA)}, 1998.

\bibitem{maniprrt}
G.~Khandate, S.~Shang, E.~T. Chang, T.~L. Saidi, Y.~Liu, S.~M. Dennis,
  J.~Adams, and M.~Ciocarlie, ``Sampling-based exploration for reinforcement
  learning of dexterous manipulation,'' in \emph{Robotics: Science and Systems
  (RSS)}, 2023.

\bibitem{unidexfpm}
T.~Wu, Y.~Gan, M.~Wu, J.~Cheng, Y.~Yang, Y.~Zhu, and H.~Dong, ``Dexterous
  functional pre-grasp manipulation with diffusion policy,'' \emph{arXiv
  preprint arXiv:2403.12421}, 2024.

\bibitem{unidexgraspplusplus}
W.~Wan, H.~Geng, Y.~Liu, Z.~Shan, Y.~Yang, L.~Yi, and H.~Wang,
  ``{UniDexGrasp++}: Improving dexterous grasping policy learning via
  geometry-aware curriculum and iterative generalist-specialist learning,'' in
  \emph{Proceedings of the IEEE/CVF International Conference on Computer Vision
  (ICCV)}, 2023.

\bibitem{unidexgrasp}
Y.~Xu, W.~Wan, J.~Zhang, H.~Liu, Z.~Shan, H.~Shen, R.~Wang, H.~Geng, Y.~Weng,
  J.~Chen, T.~Liu, L.~Yi, and H.~Wang, ``{UniDexGrasp}: Universal robotic
  dexterous grasping via learning diverse proposal generation and
  goal-conditioned policy,'' \emph{arXiv preprint arXiv:2303.00938}, 2023.

\bibitem{anydexgrasp}
H.-S. Fang, H.~Yan, Z.~Tang, H.~Fang, C.~Wang, and C.~Lu, ``{AnyDexGrasp}:
  General dexterous grasping for different hands with human-level learning
  efficiency,'' \emph{arXiv preprint arXiv:2502.16420}, 2025.

\bibitem{dexpbt}
A.~Petrenko, A.~Allshire, G.~State, A.~Handa, and V.~Makoviychuk, ``{DexPBT}:
  Scaling up dexterous manipulation for hand-arm systems with population based
  training,'' in \emph{Proceedings of Robotics: Science and Systems}, 2023.

\bibitem{dextrack}
X.~Liu, J.~Adalibieke, Q.~Han, Y.~Qin, and L.~Yi, ``{DexTrack}: Towards
  generalizable neural tracking control for dexterous manipulation from human
  references,'' in \emph{International Conference on Learning Representations
  (ICLR)}, 2025.

\bibitem{dexrepnet}
Q.~Liu, Y.~Cui, Q.~Ye, Z.~Sun, H.~Li, G.~Li, L.~Shao, and J.~Chen,
  ``{DexRepNet}: Learning dexterous robotic grasping network with geometric and
  spatial hand-object representations,'' \emph{arXiv preprint
  arXiv:2303.09806}, 2023.

\bibitem{dexrepnetpp}
Q.~Liu, Z.~Sun, Y.~Cui, H.~Li, G.~Li, L.~Shao, J.~Chen, and Q.~Ye,
  ``{DexRepNet++}: Learning dexterous robotic manipulation with geometric and
  spatial hand-object representations,'' \emph{arXiv preprint
  arXiv:2602.21811}, 2026.

\bibitem{dexhanddiff}
Z.~Liang, Y.~Mu, Y.~Wang, T.~Chen, W.~Shao, W.~Zhan, M.~Tomizuka, P.~Luo, and
  M.~Ding, ``{DexHandDiff}: Interaction-aware diffusion planning for adaptive
  dexterous manipulation,'' in \emph{Proceedings of the IEEE/CVF Conference on
  Computer Vision and Pattern Recognition (CVPR)}, 2025, pp. 1745--1755.

\bibitem{flowhoi}
H.~Zeng, L.~Chen, J.~Yang, Y.~Zhang, F.~Shi, P.~Liu, and X.~Zuo, ``{FlowHOI}:
  Flow-based semantics-grounded generation of hand-object interactions for
  dexterous robot manipulation,'' \emph{arXiv preprint arXiv:2602.13444}, 2026.

\bibitem{pkda}
W.~Bai, Q.~Chen, X.~Lin, J.~Li, Q.~Li, H.~Pan, and Y.~Sun, ``Dexterous
  manipulation transfer via progressive kinematic-dynamic alignment,'' in
  \emph{Proceedings of the AAAI Conference on Artificial Intelligence},
  vol.~40, no.~22, 2026, pp. 18\,126--18\,134.

\bibitem{teledexter}
P.~Li, Z.~Chen, Y.~Wu, P.~Wei, Y.~Li, T.~Wang, J.~Shi, M.~Yu, B.~Jia, S.-C.
  Zhu, T.~Liu, and S.~Huang, ``Towards human-level dexterous teleoperation,''
  \emph{arXiv preprint arXiv:2607.11481}, 2026.

\bibitem{omni6dpose}
J.~Zhang, W.~Huang, B.~Peng, M.~Wu, F.~Hu, Z.~Chen, B.~Zhao, and H.~Dong,
  ``{Omni6DPose}: A benchmark and model for universal {6D} object pose
  estimation and tracking,'' in \emph{European Conference on Computer Vision
  (ECCV)}, 2024, pp. 199--216.

\bibitem{bodex}
J.~Chen, Y.~Ke, and H.~Wang, ``{BODex}: Scalable and efficient robotic
  dexterous grasp synthesis using bilevel optimization,'' in \emph{IEEE
  International Conference on Robotics and Automation (ICRA)}, 2025.

\bibitem{isaacgym}
V.~Makoviychuk, L.~Wawrzyniak, Y.~Guo, M.~Lu, K.~Storey, M.~Macklin,
  D.~Hoeller, N.~Rudin, A.~Allshire, A.~Handa, and G.~State, ``{Isaac Gym}:
  High performance {GPU}-based physics simulation for robot learning,'' in
  \emph{Advances in Neural Information Processing Systems (NeurIPS) Datasets
  and Benchmarks Track}, 2021.

\bibitem{epo}
J.~Wang, Y.~Su, A.~Gupta, and D.~Pathak, ``Evolutionary policy optimization,''
  \emph{arXiv preprint arXiv:2503.19037}, 2025.

\bibitem{bps}
S.~Prokudin, C.~Lassner, and J.~Romero, ``Efficient learning on point clouds
  with basis point sets,'' \emph{arXiv preprint arXiv:1908.09186}, 2019.

\bibitem{ppo}
J.~Schulman, F.~Wolski, P.~Dhariwal, A.~Radford, and O.~Klimov, ``Proximal
  policy optimization algorithms,'' \emph{arXiv preprint arXiv:1707.06347},
  2017.

\bibitem{asymmetric_ac}
L.~Pinto, M.~Andrychowicz, P.~Welinder, W.~Zaremba, and P.~Abbeel, ``Asymmetric
  actor critic for image-based robot learning,'' in \emph{Robotics: Science and
  Systems (RSS)}, 2018.

\bibitem{foundationpose}
B.~Wen, W.~Yang, J.~Kautz, and S.~Birchfield, ``{FoundationPose}: Unified {6D}
  pose estimation and tracking of novel objects,'' in \emph{Proceedings of the
  IEEE/CVF Conference on Computer Vision and Pattern Recognition (CVPR)}, 2024,
  pp. 17\,868--17\,879.

\bibitem{sam3d}
X.~Chen, F.-J. Chu, P.~Gleize, K.~J. Liang, A.~Sax, H.~Tang, W.~Wang, M.~Guo,
  T.~Hardin, X.~Li, A.~Lin, J.~Liu, Z.~Ma, A.~Sagar, B.~Song, X.~Wang, J.~Yang,
  B.~Zhang, P.~Doll{\'a}r, G.~Gkioxari, M.~Feiszli, and J.~Malik, ``{SAM 3D}:
  {3Dfy} anything in images,'' \emph{arXiv preprint arXiv:2511.16624}, 2025.

\end{thebibliography}
